\pdfoutput=1

\documentclass[11pt]{article}

\usepackage[]{ACL2023}

\usepackage{times}
\usepackage{latexsym}
\usepackage{graphicx} 
\usepackage{multirow}
\usepackage{ulem}
\usepackage{enumitem}

\usepackage{color}

\usepackage{pifont} 

\usepackage{tcolorbox}
\tcbuselibrary{listings, breakable}

 \usepackage{soul}

\usepackage[T1]{fontenc}

\usepackage[utf8]{inputenc}
\usepackage{microtype}

\usepackage{inconsolata}

\title{RDRec: Rationale Distillation for LLM-based Recommendation }

\author{
  Xinfeng Wang$^{\dagger}$, Jin Cui$^{\dagger}$, Yoshimi Suzuki$^{\ddagger}$, \and Fumiyo Fukumoto$^{\ddagger}$ \\
  $^{\dagger}$Graduate School of Engineering \\
  $^{\ddagger}$Interdisciplinary Graduate School\\
  University of Yamanashi, Kofu, Japan \\
  \texttt{\{g22dtsa7, g22dtsa5, ysuzuki, fukumoto\}@yamanashi.ac.jp}
}

\begin{document}
\maketitle
\begin{abstract}


Large language model (LLM)-based recommender models that bridge users and items through textual prompts for effective semantic reasoning have gained considerable attention. However, few methods consider the underlying rationales behind interactions, such as user preferences and item attributes, limiting the reasoning capability of LLMs for recommendations. This paper proposes a rationale distillation recommender (RDRec), a compact model designed to learn rationales generated by a larger language model (LM). 
By leveraging rationales from reviews related to users and items, RDRec remarkably specifies their profiles for recommendations. Experiments show that RDRec achieves state-of-the-art (SOTA) performance in both top-N and sequential recommendations. 
Our source code is released at \href{https://github.com/WangXFng/RDRec}{https://github.com/WangXFng/RDRec}.
\end{abstract}


\section{Introduction}
Large language models (LLMs) with powerful reasoning capabilities have been extensively studied for recommendations, including news and item recommendations \cite{li2022miner, wei2023llmrec, huang2023recommender}, explainable recommendations \cite{yang2023large, cheng2023explainable}, and zero-/few-shot and cold-start recommendations \cite{he2023large, sanner2023large}. Several attempts have leveraged knowledge of LLMs to improve recommendation performance, such as enhancing embedding initialization \cite{harte2023leveraging}, reranking candidates \cite{yue2023llamarec}, and learning representation \cite{ren2023representation, lin2023multi, lei2023recexplainer, viswanathan2023datafinder}. 
A straightforward approach is to integrate user and item IDs into LMs through prompt learning \cite{liu2023pre}, including discrete prompts to find alternative words to represent IDs, continuous prompts to directly feed ID vectors into a pre-trained model \cite{sun2019bert4rec}, and hybrid prompts \cite{li2023personalized, zhangprompt}.
Recently, \citet{geng2022recommendation} present a P5 paradigm to transform user--item interactions, user sequential behaviors, and reviews into text-to-text prompts for LLMs. This enables P5 to capture deeper semantics for LLM-based recommendations. \citet{li2023prompt} enhance P5 by a prompt distillation, resulting in significant improvement and reductions in inference time.



\begin{figure}[t]
\begin{center}
\includegraphics[width=\linewidth]{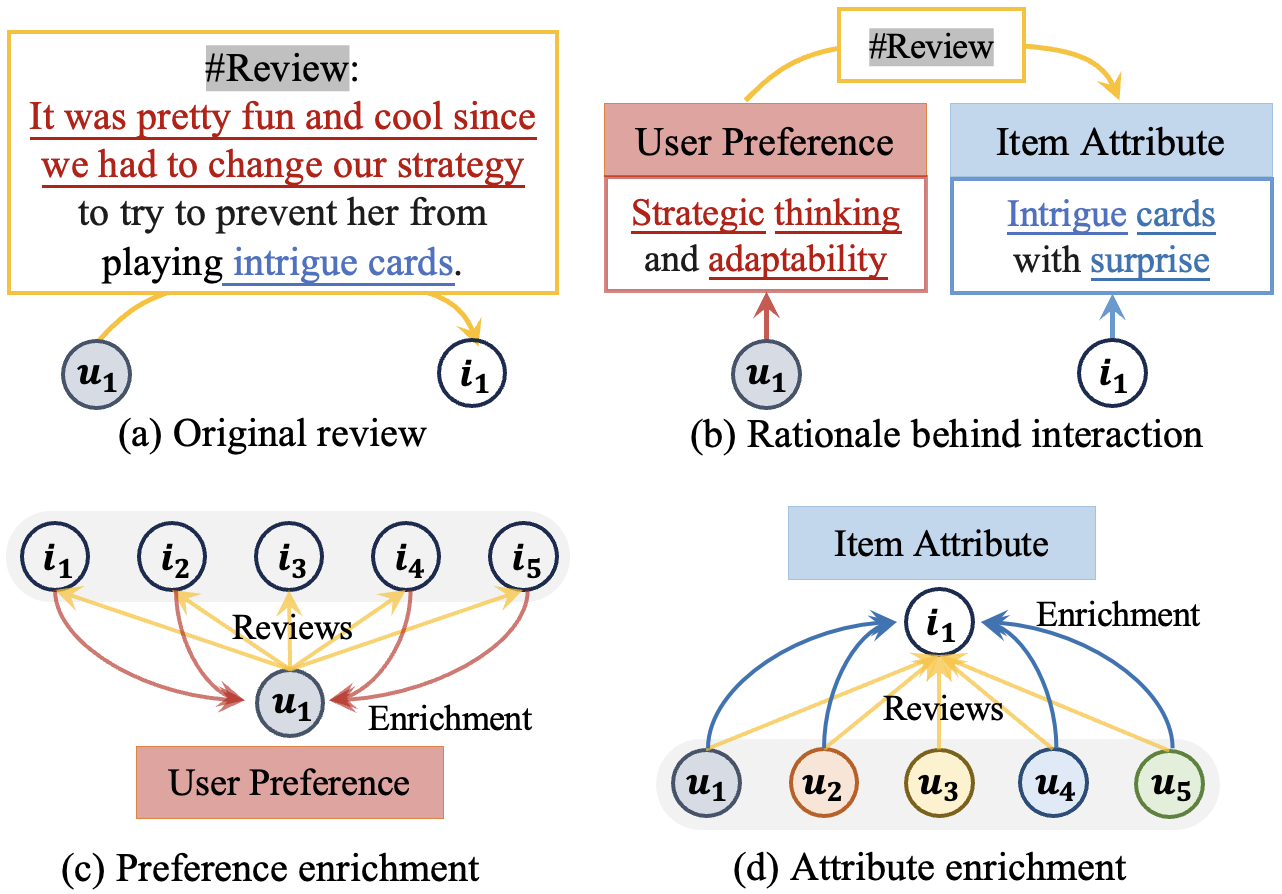} 
\caption{Illustration of our motivation. (a) denotes the review after a purchase and (b) refers to the rationale of the purchase distilled by LLMs. (c) and (d) indicate the preference and attribute enrichment, respectively.}  

\label{motivation}
\end{center}
\end{figure}

However, they pay no attention to mining the rationale behind each interaction, such as user preferences and item attributes, which hampers the reasoning capabilities of LLMs. As an example, in Fig. \ref{motivation} (a), a user review for an item says: ``\textit{\ul{It was pretty fun and cool since we had to change our strategy} \textcolor{magenta}{(user preference)} to try to prevent her from playing \ul{intrigue cards} \textcolor{blue}{(item attributes)}.}'' The user prefers strategic thinking in the game, and intrigue cards symbolize item characteristics. This introduces noise into the user's profile, as the user leans towards a strategic game rather than merely cards. 
This suggests that the original review without intermediate prompts prevents the model from learning to understand the rationale behind the interaction.

\begin{figure*}[t]
\includegraphics[width=\linewidth]{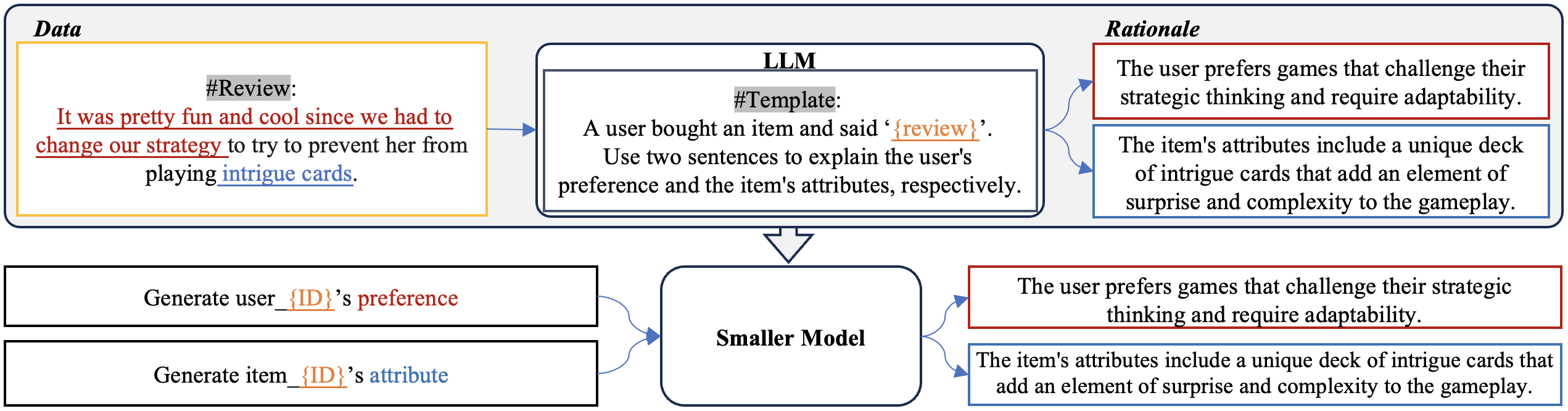} 
\caption{\label{fig:distillation}Illustration of rationale distillation with LLMs via the chain-of-thought (CoT) prompting. } 
\end{figure*}

The Chain-of-Thought (CoT) prompting \cite{wei2022chain, wang2023towards} that promises LLMs to decompose intermediate rationales, has been widely applied for rationale extraction \cite{hao2023reasoning, zhang2023recommendation, mckee2023language, zhu2023explain}. More recently, \citet{hsieh2023distilling} utilize the CoT prompting to distill rationales via LLMs to train smaller models. Inspired by this, we propose a compact recommender model to learn the interaction rationales, i.e., user preferences and item attributes, distilled from reviews using a larger LM. In this way, the model acquires clear textual knowledge with less noise (e.g., ``intrigue cards'' that may hinder understanding the user's preference for ``strategic games'' in Fig. \ref{motivation} (b)). 
This enables the model to derive more specified user and item profiles from all reviews given by the user or regarding the item for recommendations, as illustrated in Fig. \ref{motivation} (cd).




The main contributions of this paper can be summarized as follows. (1) We propose a compact RDRec model that effectively specifies user and item profiles by distilling interaction rationales from relevant reviews using a larger LM, and (2) RDRec consistently outperforms SOTA baselines on three real-world datasets in both sequential and top-N recommendations.

\section{RDRec Framework}

We present an RDRec model consisting of two stages, an interaction rationale distillation and a rationale-aware recommendation.


\begin{figure*}[t]
\begin{center}
\includegraphics[width=\linewidth]{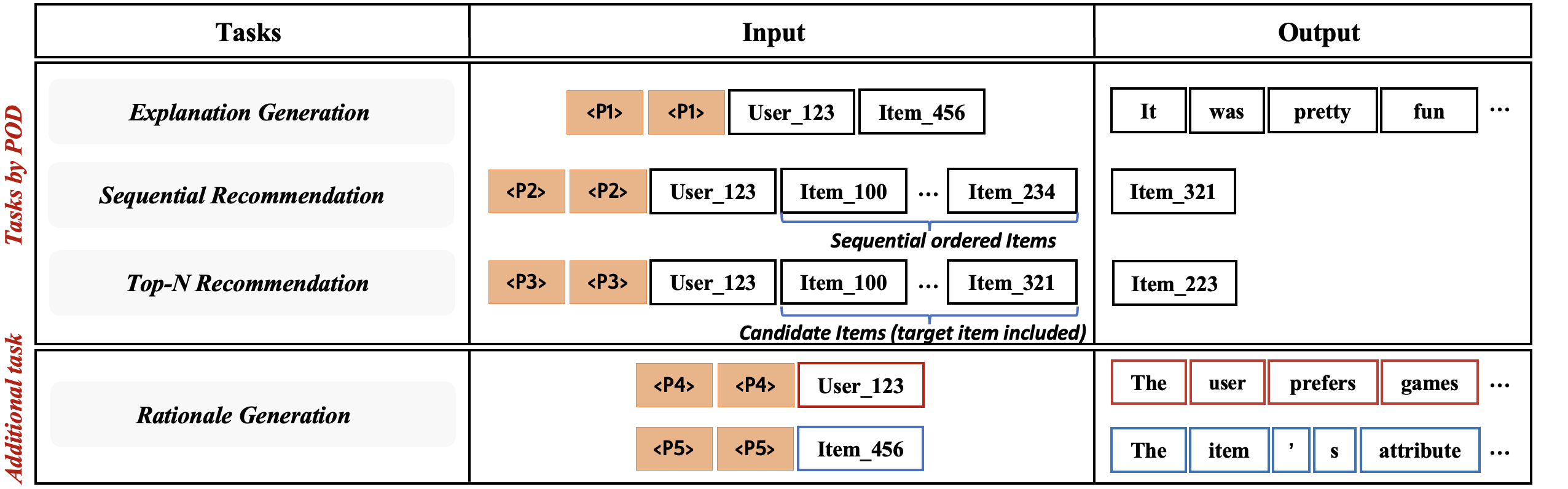} 
\caption{\label{various_tasks} Illustration of input and output of four tasks by RDRec in the prompt distillation setting. }  
\end{center}
\end{figure*}

\subsection{Interaction Rationale Distillation}

Inspired by the recent works \cite{hsieh2023distilling, miao2023exploring} that employ LLMs to produce training data for smaller models, we distill user preferences and item attributes from reviews by using the following prompt template:
``\textit{A user bought an item and said `\ul{\{$review$\}'}. Use two sentences to explain the user's preference and the item's attributes, respectively.}'' As illustrated in Fig. \ref{fig:distillation}, a review feeds into LLMs with the prompt template. The output is user preferences and item attributes. 


Formally, given a user--item interaction triplet $(u, i, r_{u,i})$ where $u$, $i$, and $r_{u,i}$ indicate a user, an item, and a review, respectively, we generate a quadruplet $(u, i, p_{u,i}, a_{u,i})$ through rationale distillation. Here, $p_{u,i}$ and $a_{u,i}$ refer to the distilled user preference and item attribute, respectively.

\subsection{Rationale-aware Recommendation}
The RDRec uses PrOmpt Distillation (POD) \cite{li2023prompt} as its backbone. 
POD converts three recommendation tasks into LLM-based text generation tasks, and then distills continuous prompt vectors from task templates. These tasks are (i) sequential recommendations, predicting the next item through the user's ordered interactions, (ii) top-N recommendations, recommending the top N items not yet engaged with by the user, and (iii) explanation generation for a user's interactions.


 In contrast to POD, we incorporate an additional rationale generation task, consisting of a user preference generation and an item attribute generation. Specifically, following POD, we first distill prompt vectors (``<P4>'' and ``<P5>'' in Fig. \ref{various_tasks}) from the templates of ``\textit{Generate user\_\ul{\{\#$u$\}}'s preference}'' and ``\textit{Generate item\_\ul{\{\#$i$\}}'s attribute}'', where \#$u$ and \#$i$ denote the user and item IDs. Then, we concatenate prompt vectors with user and item IDs as the input, and the generated preference $p_{u,i}$ and attribute $a_{u,i}$ as the output to train the model. To address the token composing issue (i.e., the token of ``user\_123'' is often tokenized by LLMs as a sequence of [``user'', ``\_'', ``12'' and ``3'']), we use the whole-word embedding \cite{geng2022recommendation} to treat each sequence of ID tokens as a complete unit, making it distinguishable as a word. 

Fig. \ref{various_tasks} illustrates the input and output example of the four tasks. 
We define a pair of input-output words as $X$ = [$x_1$, $...$, $x_{|X|}$] and $Y$ = [$y_1$, $...$, $y_{|Y|}$], respectively. We then concatenate the tokens of the input with prompt vectors and obtain [$\mathbf{x}_1$, $...$, $ \mathbf{x}_{|X|}$, $\mathbf{p}_1$, $...$, $\mathbf{p}_{|P|}$]. After adding the whole-word representation [$\mathbf{w}_1$, $...$, $ \mathbf{w}_{|X|+|P|}$], we feed them into the smaller model in RDRec to obtain a probability distribution $p(y|Y_{<t}, X)$ over a vocabulary at each step $t$, where $Y_{<t}$ denotes the tokens generated before step $t$. We adopt a log-likelihood loss function to optimize the model parameters $\Theta$:
\begin{equation}
\label{eq:sa}
\mathcal{L}_{\Theta} = \frac{1}{|\mathcal{D}|} \sum_{(X,Y) \in \mathcal{D}} \frac{1}{|Y|}\sum_{t=1}^{|Y|} -\log p(y|Y_{<t}, X),
\end{equation}

\noindent
where $\mathcal{D}$ denotes the training set consisting of all input-output pairs for four tasks. $|\mathcal{D}|$ and $|Y|$ denote the amount of training samples and the number of tokens in the output sequence, respectively.

\subsection{Model Optimization and Inference}
Following POD, we shuffle the input-output pairs of four tasks and randomly select samples from each task in a specified proportion. 
We thereafter mixed these samples to train the RDRec model.
During inference, we employ a beam search algorithm to generate results by selecting the word with the highest likelihood from the vocabulary.

\section{Experiment}
\subsection{Experimental Setup}
\textbf{Datasets and Metrics.} Consistent with POD, we performed experiments on three public datasets, i.e.,  Sports \& Outdoors, Beauty, and Toys \& Games, which are collected from the Amazon dataset\footnote{https://www.amazon.com/}. 
For direct and sequential recommendation tasks, we designate the last interaction as the test label, the second-to-last as the validation label, and the remaining interactions as training data. The statistics of datasets are provided in Table \ref{tab:statistics}. To evaluate the recommendation performance, we utilized the evaluation metrics of hit rate (HR)@$k$ (H@$k$) and normalized discounted cumulative gain (NDCG)@$k$ (N@$k$) with $k \in \{1, 5, 10\}$. 

\noindent
\textbf{Baselines.} We compared RDRec with ten baselines for sequential recommendations: CASER \cite{tang2018personalized}, HGN \cite{ma2019hierarchical}, GRU4Rec \cite{hidasi2015session}, BERT4Rec \cite{sun2019bert4rec}, FDSA \cite{zhang2019feature}, SASRec \cite{kang2018self},  S$^3$-Rec \cite{zhou2020s3}, P5 \cite{geng2022recommendation}, RLS \cite{chu2023leveraging} and POD \cite{li2023prompt}. We compared RDRec with five baselines for top-N recommendations: MF \cite{koren2009matrix}, MLP \cite{cheng2016wide}, P5 \cite{geng2022recommendation}, RLS \cite{chu2023leveraging} and POD \cite{li2023prompt}.

\noindent
\textbf{Implementation.} For a fair comparison,  RDRec used T5-small \cite{raffel2020exploring} as the smaller model, aligning with the baselines P5 and POD. We used Llama-2-7b \cite{touvron2023llama} as the larger LM. We reported a 10-trial T-test to show the robustness of RDRec. Our RDRec was implemented and experimented with Pytorch on Nvidia GeForce RTX 3090 (24GB memory). 
The Appendix \ref{sec:experiment_detail} provides further details. 

\begin{table}
\centering
\resizebox{\linewidth}{!}{ 
  \begin{tabular}{cccccc}
\hline
    {Dataset} &{\#User} &{\#Item} &{\#Review}& {Avg.}& Density (\%)\\
\hline
    {Sports} &48,993& 34,298& 296,337& 8.3 & 0.0453 \\
    {Beauty} & 22,363& 12,101& 198,502& 8.9 & 0.0734 \\
    {Toys} &19,804& 22,086&  167,597& 8.6 & 0.0724 \\
\hline
\end{tabular}
}
  \caption{\label{tab:statistics}Statistics of dataset. ``\#User'', ``\#Item'', ``\#Review'', and ``Avg.'' denote the number of users, items, reviews, and average user reviews, respectively.}
\end{table}

\begin{table*}
\centering
\resizebox{.86\linewidth}{!}{ 
\begin{tabular}{c|cccc|cccc|cccc}
\hline
\multirow {2}{*}{Models} & \multicolumn{4}{c|}{Sports} & \multicolumn{4}{c|}{Beauty} & \multicolumn{4}{c}{Toys}\\
& H@5 &N@5 & H@10 & N@10 & H@5 &N@5 & H@10 & N@10 & H@5 &N@5 & H@10 & N@10\\
\hline
Caser & 0.0116 & 0.0072 & 0.0194 & 0.0097 & 0.0205 & 0.0131 & 0.0347 & 0.0176 & 0.0166 & 0.0107 & 0.0270 & 0.0141\\
HGN & 0.0189 & 0.0120 & 0.0313 & 0.0159 & 0.0325 & 0.0206 & 0.0512 & 0.0266 & 0.0321 & 0.0221 & 0.0497 & 0.0277\\
GRU4Rec & 0.0129 & 0.0086 & 0.0204 & 0.0110 & 0.0164 & 0.0099 & 0.0283 & 0.0137 & 0.0097 & 0.0059 & 0.0176 & 0.0084\\
BERT4Rec & 0.0115 & 0.0075 & 0.0191 & 0.0099 & 0.0203 & 0.0124 & 0.0347 & 0.0170 & 0.0116 & 0.0071 & 0.0203 & 0.0099\\
FDSA & 0.0182 & 0.0122 & 0.0288 & 0.0156 & 0.0267 & 0.0163 & 0.0407 & 0.0208 & 0.0228 & 0.0140 & 0.0381 & 0.0189\\
SASRec & 0.0233 & 0.0154 & 0.0350 & 0.0192 & 0.0387 & 0.0249 & 0.0605 & 0.0318 & 0.0463 & 0.0306 & 0.0675 & 0.0374\\
S$^3$-Rec & 0.0251 & 0.0161 & 0.0385 & 0.0204 & 0.0387 & 0.0244 & 0.0647 & 0.0327 & 0.0443 & 0.0294 & 0.0700 & 0.0376\\
P5 & 0.0387 & 0.0312 & 0.0460 & 0.0336 & 0.0508 & 0.0379 & 0.0664 & 0.0429 & 0.0648 & 0.0567 & 0.0709 & 0.0587\\
RSL & 0.0392 & 0.0330 & 0.0512 & 0.0375 & 0.0508 & 0.0381 & 0.0667 & 0.0446 & 0.0676 & 0.0583 & 0.0712 & 0.0596\\
POD & \ul{0.0497} & \ul{0.0399} & \ul{0.0579} & \ul{0.0422} & \ul{0.0559} & \ul{0.0420} & \ul{0.0696} & \ul{0.0471} & \ul{0.0692} & \ul{0.0589} & \ul{0.0749} & \ul{0.0601} \\
\textbf{Ours} & \textbf{0.0505} & \textbf{0.0408} & \textbf{0.0596} & \textbf{0.0433} & \textbf{0.0601} & \textbf{0.0461} & \textbf{0.0743} & \textbf{0.0504} & \textbf{0.0723} & \textbf{0.0593} & \textbf{0.0802} & \textbf{0.0605} \\
\hline
Impv (\%). & \textbf{1.6} & \textbf{2.2} & \textbf{2.8} & \textbf{2.5} & \textbf{7.5}* & \textbf{9.8}* & \textbf{6.7}* & \textbf{7.1}* & \textbf{4.4}* & \textbf{0.6} & \textbf{7.1}* & \textbf{0.7} \\
p-value & 6.3e-1 & 5.1e-1 & 2.7e-1 & 3.8e-1 & 8.1e-3 & 2.4e-3 & 2.1e-2 & 2.5e-2 & 1.0e-2 & 5.8e-1 & 1.7e-5 & 5.9e-1\\
\hline
\end{tabular}
}
\caption{\label{tab:sequential_recommendation} Performance comparison on sequential recommendation. \textbf{Bold}: Best, \ul{underline}: Second best. ``*'' indicates that the improvement is statistically significant (p-value < 0.05) in the 10-trial T-test. All of the baselines are reported by the papers \cite{geng2022recommendation, chu2023leveraging, li2023prompt}, except for the POD model.}
\end{table*}

\begin{table*}
\centering
\resizebox{1\linewidth}{!}{ 
\begin{tabular}{c|ccccc|ccccc|ccccc}
\hline
\multirow {2}{*}{Models} & \multicolumn{5}{|c|}{Sports} & \multicolumn{5}{c|}{Beauty} & \multicolumn{5}{c}{Toys}\\
& H@1& H@5 &N@5 & H@10 & N@10 & H@1 & H@5 &N@5 & H@10 & N@10 & H@1 & H@5 &N@5 & H@10 & N@10\\
\hline
MF & 0.0314 & 0.1404 & 0.0848 & 0.2563 & 0.1220 & 0.0311 & 0.1426 & 0.0857 & 0.2573 & 0.1224 & 0.0233 & 0.1066 & 0.0641 & 0.2003 & 0.0940\\
MLP & 0.0351 & 0.1520 & 0.0927 & 0.2671 & 0.1296 & 0.0317 & 0.1392 & 0.0848 & 0.2542 & 0.1215 & 0.0252 & 0.1142 & 0.0688 & 0.2077 & 0.0988 \\
P5 & 0.0726 & 0.1955 & 0.1355 & 0.2802 & 0.1627 & 0.0608 & 0.1564 & 0.1096 & 0.2300 & 0.1332 & 0.0451 & 0.1322 & 0.0889 & 0.2023 & 0.1114 \\
RSL & 0.0892 & 0.2092 & 0.1502 & \ul{0.3001} & 0.1703 & 0.0607 & 0.1612 & 0.1110 & 0.2209 & 0.1302 & 0.0389 & 0.1423 & 0.0825 & 0.1926 & 0.1028 \\
POD & \ul{0.0927} & \ul{0.2105} & \ul{0.1539} & {0.2889} & \ul{0.1782} & \ul{0.0846} & \ul{0.1931} & \ul{0.1404} & \ul{0.2677} & \ul{0.1639} & \ul{0.0579} & \ul{0.1461} & \ul{0.1029} & \ul{0.2119} & \ul{0.1244}\\
\textbf{Ours} & \textbf{0.1285} & \textbf{0.2747} & \textbf{0.2033} & \textbf{0.3683} & \textbf{0.2326} & \textbf{0.1203} & \textbf{0.2572} & \textbf{0.1902} & \textbf{0.3380} & \textbf{0.2160} & \textbf{0.0660} & \textbf{0.1655} & \textbf{0.1171} & \textbf{0.2375} & \textbf{0.1398} \\
\hline
Impv. (\%) & \textbf{38.6}*& \textbf{30.5}*& \textbf{32.1}* & \textbf{27.5}* & \textbf{30.5}* & \textbf{42.2}* & \textbf{33.2}* & \textbf{35.8}* & \textbf{26.3}* & \textbf{31.8}* & \textbf{13.9}* & \textbf{13.2}* & \textbf{13.8}* & \textbf{12.1}* & \textbf{12.4}* \\
p-value & 2.3e-14 & 1.1e-14 & 2.8e-15 & 1.1e-16 & 5.0e-15 & 3.8e-15 & 2.0e-15 & 1.7e-15 & 2.7e-15 & 2.1e-15  & 5.6e-7 & 4.4e-8 & 2.4e-8 & 1.2e-8 & 9.8e-9 \\
\hline
\end{tabular}
}
\caption{\label{tab:top_n_recommendation} Comparison on top-N recommendation. The T-test shows the results by RDRec and the second-best, POD.}


\end{table*}

\subsection{Experimental Results}




Tables \ref{tab:sequential_recommendation} and \ref{tab:top_n_recommendation} show comparative results between RDRec and baselines. We can see that the RDRec consistently surpasses the runner-ups, POD and RSL, with the improvement of 0.5 $\sim$ 9.8\% in H@$k$ and N@$k$ for sequential recommendations, and 12.1 $\sim$ 42.2\% in H@$k$ and N@$k$ for top-N recommendations, where $k \in \{1, 5, 10\}$. This highlights the effectiveness of learning interaction rationales to improve both recommendation tasks. 

We also observed that RDRec exhibits greater improvement in top-N recommendations compared to sequential recommendations. 
This indicates that specifying user preferences and item attributes is more beneficial to recommending top-N unknown candidates, whereas sequential recommenders rely more on capturing correct behavioral patterns for predicting the user's next choice.

%


\begin{table}
\centering
\resizebox{.95\linewidth}{!}{ 
  \begin{tabular}{cc|cc|cc|cc}
\hline
    \multirow {2}{*}{UsP} &  \multirow {2}{*}{ItA} & \multicolumn{2}{|c|}{Sports} & \multicolumn{2}{c|}{Beauty} & \multicolumn{2}{c}{Toys}\\
 & & H@10 & N@10 & H@10 &N@10 & H@10 & N@10\\
\hline
\multicolumn{8}{c}{Sequential recommendation}\\
\hline
   \ding{56} & \ding{56} & {0.0566} & {0.0408} & {0.0705} & {0.0479} & {0.0768} & {0.0573} \\
   \ding{52} & \ding{56} & \ul{0.0581} & \ul{0.0425} & \ul{0.0729} & \ul{0.0494} & {0.0787} & 0.0589\\
   \ding{56} & \ding{52}&  {0.0573} & {0.0411} & {0.0712} & {0.0492} & \ul{0.0788} & \ul{0.0593}\\
   \ding{52} & \ding{52}& \textbf{0.0596} & \textbf{0.0433} & \textbf{0.0743} & \textbf{0.0504} & \textbf{0.0802} & \textbf{0.0605} \\
\hline
\multicolumn{8}{c}{Top-N recommendation}\\
\hline
    \ding{56} & \ding{56}  & 0.2977 & 0.1850 & 0.2777 & 0.1701 & 0.2200 & 0.1284 \\
    \ding{52} & \ding{56} & 0.3509 & 0.2200 & 0.3080 & 0.1912 & 0.2214 & 0.1307 \\
    \ding{56} & \ding{52}& \ul{0.3513} & \ul{0.2249} & \ul{0.3275} & \ul{0.2048} & \ul{0.2321} & \ul{0.1370} \\
    \ding{52} & \ding{52}& \textbf{0.3683} & \textbf{0.2326} & \textbf{0.3380} & \textbf{0.2160} & \textbf{0.2375} & \textbf{0.1398}\\
\hline
\end{tabular}
    }
  \caption{\label{tab:ablation_study} Ablation study. ``w/o X'' denotes the removed parts. ``UsP'' and ``ItA'' indicate the distillation of user preferences and item attributes, respectively.}
\end{table}

We conducted an ablation experiment to examine the rationale distillation. The result in Table. \ref{tab:ablation_study} shows that distilling user preferences and item attributes from reviews is advantageous for both sequential and top-N recommendations. We can see that specifying item profiles is generally more effective for top-N recommendation, whereas specifying user profiles is more effective for sequential recommendation on the Sports and Beauty datasets. 

\subsection{Error Analysis of Sequential Recommendation}

We conducted an error analysis to examine the sequential recommendations by RDRec. We identified two noteworthy error cases:

\noindent
\textbf{Case (i)}. RDRec may prioritize the next item based on a user's earlier interactions rather than recent ones. One reason is that the Transformer \cite{vaswani2017attention} in T5 excels in capturing long-term dependencies, while it may cause RDRec to pay less attention to recent interactions. This suggests to enhance its self-attention \cite{fan2022sequential} or develop short-term prompt-aware templates for LLM-based sequential recommendations. 

\noindent
\textbf{Case (ii)}. RDRec often disregards popular items for users because they do not align with their sequential patterns. One possible reason for this is that, during training, RDRec selects random subsequences from the user interaction sequence and predicts the last item of each subsequence. This process emphasizes sequential patterns but possibly sacrifices the model's capability to identify popular items. This suggests introducing a popularity-based interaction graph to help the model be aware of popular high-order neighbors.

\subsection{In-Depth Analysis of RDRec}
To better understand the RDRec, we conducted in-depth experiments and analysis. The Appendix \ref{sec:discussion} provides further analyses. 
\begin{table}
\centering
\resizebox{\linewidth}{!}{ 
  \begin{tabular}{c|cc|cc|cc}
\hline
    Ratio & \multicolumn{2}{|c|}{Sports} & \multicolumn{2}{c|}{Beauty} & \multicolumn{2}{c}{Toys}\\
 EG:RG:SR:TR& H@10 & N@10 & H@10 &N@10 & H@10 & N@10\\
\hline
\multicolumn{7}{c}{Sequential recommendation}\\
\hline
    1\ :\ 1\ :\ 1\ :\ 1 & \textbf{0.0596} & \textbf{0.0433} & \textbf{0.0743} & \textbf{0.0504} & 0.0789 & 0.0594\\
    1\ :\ 1\ :\ 2\ :\ 1 & \ul{0.0593} & \ul{0.0431} & \ul{0.0735} & \ul{0.0502} & \ul{0.0790} & \ul{0.0601} \\
    1\ :\ 1\ :\ 1\ :\ 3 & 0.0592 & 0.0426 & 0.0702 & 0.0445 & \textbf{0.0802} & \textbf{0.0605} \\
\hline
\multicolumn{7}{c}{Top-N recommendation}\\
\hline
    1\ :\ 1\ :\ 1\ :\ 1 & \ul{0.3261} & \ul{0.2022} & \ul{0.2855} & \ul{0.1854} & \ul{0.2214} & \ul{0.1307} \\
    1\ :\ 1\ :\ 2\ :\ 1 & 0.2822 & 0.1722 & 0.2693 & 0.1584 & 0.1872 & 0.1037 \\
    1\ :\ 1\ :\ 1\ :\ 3 & \textbf{0.3683} & \textbf{0.2160} & \textbf{0.3380} & \textbf{0.2160} & \textbf{0.2375} & \textbf{0.1398}\\
\hline
\end{tabular}
    }
  \caption{\label{tab:ratio} Performance on the sample ratios of various tasks. ``EG'', ``RG'', ``SR'' and ``TR'' denote explanation and rationale generation, and sequential and top-N recommendations, respectively.}
\end{table}

\noindent
\textbf{Effect of sample ratios}. 
We observe from Table \ref{tab:ratio} that on the Toys dataset, increasing the ratio of top-N samples for training RDRec improves sequential recommendations, while in major cases a higher ratio of sequential samples always harms top-N recommendations.
One reason is that the training strategy of sequential tasks prioritizes sequential patterns while compromising its ability to detect unknown items.



%



\noindent
\textbf{Computational complexity}. 
Both Llama2 and T5 are Transformer-based models, with computational complexity of $\mathcal{O}(L^2)$, where $L$ is the number of word tokens. Consequently, RDRec's computational complexity relies on user interaction count rather than the number of users and items. Compared with other complex ID-based methods, such as graph convolution network-based approaches with $\mathcal{O}((M+N)^2)$ \cite{he2020lightgcn, yu2022graph, wang2023eedn, wang2024nfarec}, where $M$ and $N$ are the numbers of users and items, respectively, and ($M$+$N$) $\gg$ $L$ in Table \ref{tab:statistics}, RDRec exhibits reduced computational demands, thereby rendering it suitable for deployment in large-scale applications.


\noindent
\textbf{Study of rationale distillation}. 
We investigated the rationale distillation and obtained two findings. One is that, even when a user negatively reviews an item, the LLM objectively specifies user requirements and item attributes. For instance, in the following input, the customer advises not buying the book unless the kids are truly interested in it. However, many others provide positive comments, such as ``\ul{\textit{The toy was really nice.}}'' and ``\ul{\textit{Fun little toy to match the book}}.''

\begin{tcolorbox}[breakable, colback=white, left=2pt, right=2pt]
\small
\textbf{Input:} 
\\ 
\textcolor{black}{\textit{\ul{My Nephew is all about trucks and machines it's cute for him but unless the kid's really into the book or just general construction I wouldn't bother.}}}
\end{tcolorbox}

\noindent
 This indicates that objective profiles (e.g., a book and its content) are more crucial than users' subjective opinions in real-world recommendations. We found that the generated item attributes by the LLM are relatively objective which is shown as follows:

\begin{tcolorbox}[breakable, colback=white, left=2pt, right=2pt]
\small
\textbf{Output:} \\ 
\textcolor{magenta}{\textit{\ul{The user prefers items that are cute and appealing to children, but not necessarily related to construction or machines.}}} \\ \\
\textcolor{blue}{\textit{\ul{The item's attributes include being a colorful and engaging picture book that teaches children about different construction vehicles.}}}
\end{tcolorbox}

\noindent
This could be a reason for the noticeable improvement in performance by learning rationales. 

The other observation is that, when a review is extremely short, the prompt could urge the LLM to produce hallucinations during rationale distillations. Recently, \citet{zhang2023generative} have proposed to mitigate hallucinations of LLM-based recommender to enhance its performance. This is a rich space for further exploration \cite{liu2022token, gao2023rarr, peng2023check}.

%


\section{Conclusion}
We proposed a compact RDRec model to learn the underlying rationales for interactions generated by a larger LM. By learning rationales from all related reviews, RDRec effectively specifies user and item profiles for recommendations. Experimental results showed the effectiveness of our RDRec. Future work involves (i) exploring better prompts for sequential recommendations, and (ii) enhancing explanation generation in RDRec.

\section*{Acknowledgements}
We would like to thank anonymous reviewers for their thorough comments and suggestions. This work is supported by the China Scholarship Council (No.202208330093) and JKA (No.2023M-401).

\section*{Ethics Statement}
This paper does not involve the presentation of a new dataset, an NLP application, and the utilization of demographic or identity characteristics information.

\section*{Limitation}
The hallucination issue during rationale distillation remains unsolved. 
Additionally, RDRec faces an unfaithful reasoning problem, misinterpreting user opinions about candidate items despite delivering correct recommendations.

\normalem
\bibliography{anthology,custom}
\bibliographystyle{acl_natbib}

\clearpage
\appendix

\section{Appendix}
\label{sec:appendix}

\subsection{Experimental Details}
\label{sec:experiment_detail}

This section provides further experimental results and implementation setup.
We focus on the fourth task as \citet{geng2022recommendation} have conducted a thorough study for the others.

\begin{table*}
\centering
\resizebox{\linewidth}{!}{ 
\begin{tabular}{c|cccc|cccc|cccc}
\hline
Ratio & \multicolumn{4}{c|}{Sports} & \multicolumn{4}{c|}{Beauty} & \multicolumn{4}{c}{Toys}\\
EG:RG:SR:TR& H@5 &N@5 & H@10 & N@10 & H@5 &N@5 & H@10 & N@10 & H@5 &N@5 & H@10 & N@10\\
\hline
\multicolumn{13}{c}{Sequential recommendation}\\
\hline
1\ :\ 1\ :\ 1\ :\ 1 & \textbf{0.0503} & \textbf{0.0402} & \textbf{0.0596} & \textbf{0.0433} & \textbf{0.0601} & \textbf{0.0461} & \textbf{0.0743} & \textbf{0.0504} & 0.0716 & 0.0579 & 0.0789 & 0.0594\\
1\ :\ 1\ :\ 2\ :\ 1 & \ul{0.0501} & \ul{0.0398} & \ul{0.0593} & \ul{0.0431} & \ul{0.0595} & \ul{0.0457} & \ul{0.0735} & \ul{0.0502} & 0.0713 & 0.0581 & 0.0790 & 0.0601\\
1\ :\ 1\ :\ 3\ :\ 1 & 0.0496 & 0.0399 & 0.0578 & 0.0420 & 0.0573 & 0.0417 & 0.0662 & 0.0452 & 0.0713 & 0.0583 & 0.0792 & 0.0601\\
1\ :\ 1\ :\ 1\ :\ 2 & 0.0489 & 0.0374 & 0.0571 & 0.0387 & 0.0565 & 0.0419 & 0.0715 & 0.0466 & \ul{0.0717} & \ul{0.0588} & \ul{0.0799} & \ul{0.0602}\\
1\ :\ 1\ :\ 1\ :\ 3  & 0.0483 & 0.0369 & 0.0592 & 0.0426 & {0.0547} & {0.0395} & {0.0702} & {0.0445} & \textbf{0.0723} & \textbf{0.0593} & \textbf{0.0802} & \textbf{0.0605} \\
\hline
\multicolumn{13}{c}{Top-N recommendation}\\
\hline
1\ :\ 1\ :\ 1\ :\ 1 & 0.2381 & 0.1750 & 0.3261 & 0.2022 & 0.2136 & 0.1516 & 0.2885 & 0.1854 & 0.1482 & 0.1062 & 0.2144 & 0.1307\\
1\ :\ 1\ :\ 2\ :\ 1 & 0.2042 & 0.1476 & 0.2822 & 0.1722 & 0.1845 & 0.1350 & 0.2693 & 0.1584 & 0.1253 & 0.0876 & 0.1872 & 0.1037\\
1\ :\ 1\ :\ 3\ :\ 1 & 0.1524 & 0.1080 & 0.2101 & 0.1298  & 0.1424 & 0.1024 & 0.2178 & 0.1359 & 0.1118 & 0.0780 & 0.1803 & 0.0998\\
1\ :\ 1\ :\ 1\ :\ 2 & \ul{0.2439} & \ul{0.1810} & \ul{0.3303} & \ul{0.2067} & \ul{0.2372} & \ul{0.1784} & \ul{0.3237} & \ul{0.2030} & \ul{0.1579} & \ul{0.1091} & \ul{0.2221} & \ul{0.1339}\\
1\ :\ 1\ :\ 1\ :\ 3 & \textbf{0.2747} & \textbf{0.2033} & \textbf{0.3683} & \textbf{0.2326} & \textbf{0.2572} & \textbf{0.1902} & \textbf{0.3380} & \textbf{0.2160} & \textbf{0.1655} & \textbf{0.1171} & \textbf{0.2375} & \textbf{0.1398} \\
\hline

\end{tabular}
}
\caption{\label{tab:detail_of_sample_ratio} Performance comparison on various sample ratios for training RDRec. ``EG'', ``RG'', ``SR'' and ``TR'' denote explanation generation, rationale generation, and sequential and top-N recommendations, respectively.}
\end{table*}

\subsubsection{Effect of Various Sample Ratios}
Table \ref{tab:detail_of_sample_ratio} shows the effect of various sample ratios on the recommendation performance. We can see that the sample ratio of various tasks for pretraining the model would influence the recommendation performance. Specifically, on the Toys dataset, increasing the ratio of top-N samples sometimes improves both sequential and top-N recommendations. In contrast, a higher ratio of sequential samples often negatively affects the performance of top-N recommendations across all datasets. 
The reason is that during training, RDRec selects random user interaction subsequences and predicts the last item of each subsequence. This process emphasizes sequential patterns, although possibly sacrifices the model's capability to identify popular items.


\subsubsection{Execution Time}
Table \ref{tab:running_time} shows the execution time in various stages by RDRec on three datasets. These results were obtained through Nvidia GeForce RTX 3090 (24GB memory). We can see that RDRec efficiently makes inferences for recommendations with a small backbone, while the interaction rationale distillation and pre-training are time-consuming. Fortunately, these processes are only required once.

\begin{table}
\centering
\resizebox{\linewidth}{!}{ 
  \begin{tabular}{c|cccc}
\hline
    \multirow {2}{*}{\textbf{Datasets}} & \multicolumn{4}{c}{\textbf{Stages}} \\
    & \textbf{Distillation} & \textbf{Pre-training} & \textbf{SR} & \textbf{TR}\\
\hline
    Sports & 16h46m28s  & 16h23m23s & 15m03s & 18m23s \\
\hline
    Beauty & 11h50m14s & 12h45m12s & 13m33s & 16m07s \\
\hline
    Toys & 09h13m05s & 08h39m37s & 16m25s & 18m21s \\
\hline
\end{tabular}
}
\caption{\label{tab:running_time} Execution time in various stages. ``SR'' and ``TR'' represent the cumulative inference time for all users in sequential and top-N recommendations, respectively. ``h'', ``m'', and ``s'' refer to ``hours'' and ``minutes'', and ``seconds'' respectively.}
\end{table}

\subsubsection{Implementation Details}
For a fair comparison, all the hyperparameters of RDRec are in the same setting as POD. Specifically, both the encoder and decoder consist of 6 layers with each layer comprising an 8-headed attention layer. The vocabulary of T5 contains a total number of 32,100 tokens, with an embedding dimensionality of 512. We iteratively and randomly sampled a segment from a user's item sequence for training the sequential recommendation task. The number of negative items for top-N recommendation is set to 99 for both training and evaluation. We used the AdamW optimizer \cite{loshchilov2017decoupled}. We set the number of prompt vectors to 3 for all tasks, the batch size for training all three tasks to 64, and the learning rate to 0.001 for the Sports dataset and 0.0005 for both the Beauty and Toys datasets. 
We split each dataset into training, validation, and test sets with a ratio of 8:1:1 for explanation generation. 
We exploit the discrete prompt templates for different tasks from \cite{geng2022recommendation}.
During training, we save a checkpoint if the total validation loss of the model in all tasks is the lowest for the current epoch. If this doesn't occur 5 times, we terminate the training process and load the best checkpoint for evaluation. At the inference stage, we set the number of beams at 20 for sequential and top-N recommendations. 


\subsubsection{Baselines}
To evaluate the performance of sequential and top-N recommendations, we compared our RDRec with twelve baselines: 

\begin{itemize}[nosep,labelindent=0em,leftmargin=1em,font=\normalfont]

\item \textbf{MF} \cite{koren2009matrix} accesses the inner product between user and item latent factors for predicting users' preference for candidates.

\item \textbf{GRU4Rec} \cite{hidasi2015session} regards the entire item sequence of each user as the user's session to recommend.

\item \textbf{MLP} \cite{cheng2016wide} exploits a stack of non-linear layers to learn user and item embeddings for making recommendations.

\item \textbf{CASER} \cite{tang2018personalized} treats user interactions as images and employs 2-dimensional convolutions to capture sequential patterns.

\item \textbf{SASRec} \cite{kang2018self} exploits Markov Chains to excavate short-term semantics in users' sequential patterns.

\item \textbf{HGN}  \cite{ma2019hierarchical} exploits a novel gating strategy to model users' long- and short-term interests in candidate items.

\item \textbf{BERT4Rec} \cite{sun2019bert4rec} proposes to leverage the BERT-style cloze task for the sequential recommender algorithm.

\item \textbf{FDSA} \cite{zhang2019feature} incorporates item features with item sequences of users to perform recommendations.

\item \textbf{S$^3$-Rec} \cite{zhou2020s3} learns users' latent behavioral features via employing a self-supervised learning paradigm.

\item \textbf{P5} \cite{geng2022recommendation} converts three different recommendation tasks into textual generation tasks using LLMs for recommendations.


\item \textbf{RSL} \cite{chu2023leveraging} adopts novel training and inference strategies to deliver LLM-based recommendations.

\item \textbf{POD} \cite{li2023prompt} refines P5 through prompt distillation to make efficient and precise recommendations.

\end{itemize}

\subsection{Further Analyses}
\label{sec:discussion}

\subsubsection{Hallucination by LLMs}

We observed that when a review is too short, the LLM might produce hallucinations. The following example illustrates a rationale with hallucinations generated by the LLM:
\begin{tcolorbox}[breakable, colback=white, left=2pt, right=2pt]
\small
\textbf{Input:} 
\\ 
\textcolor{black}{\textit{\ul{This is a fantastic game.}}}
\\

\textbf{Output:} \\ 
\textcolor{magenta}{\textit{\ul{The user prefers games with}}} \colorbox{lightgray}{engaging storylines}. \\ \\
\textcolor{blue}{\textit{\ul{The item's attributes include}}} \colorbox{lightgray}{a realistic game world}, \colorbox{lightgray}{immersive sound effects}, and \colorbox{lightgray}{smooth gameplay}.
\end{tcolorbox}

The contents of ``\ul{\textit{engaging storylines}}'', ``\ul{\textit{immersive sound effects}}'' and ``\ul{\textit{smooth gameplay}}'' marked by \colorbox{lightgray}{gray} are hallucinations overly inferred by the LLM. Toward this, mitigating hallucinations of LLM-based recommender is a rich space for future exploration \cite{liu2022token, gao2023rarr, peng2023check,zhang2023generative}.

\subsubsection{Effect of Explanation Generation}
We observed that RDRec can generate correct explanations in many cases, such as the explanation ``\ul{\textit{This is a great product for the price}},'' for the provided review ``\ul{\textit{very good quality for the price}}.''  
%
 

However, RDRec sometimes recommends candidates correctly but provides explanations that completely differ from the user's review. For instance, the generated explanation is, ``\ul{\textit{Absolutely great product}},'' whereas the user's actual review is, ``\ul{\textit{I wouldn't recommend this for painting your full nail}}.'' One possible reason is that RDRec has learned to prioritize predicting user-item interaction over considering the rationale for making recommendations. This is a challenging yet intriguing path to further improve RDRec.

\subsubsection{The Whole-Word Embedding}
To address the token composing issue (i.e., the token of ``user\_1234'' is often tokenized by the tokenizer of LLMs as a sequence of [``user'', ``\_'', ``12'' and ``34'']), we employed the whole-word embedding \cite{geng2022recommendation} to ensure that each sequence of ID tokens is a complete unit and can be distinguished from a word.


It is noteworthy that the whole-word embedding will not cause scalability issues because we only need to identify which tokens represent the same user (or item). For instance, given a token list [``P1'', ``P2'', ``P3'', ``user'', ``\_'', ``12'', ``34'', ``item'', ``\_'', ``98'', ``76''], the index list over the whole-word embedding vocabulary is [0, 0, 0, 1, 1, 1, 1, 2, 2, 2, 2]. Since the number of negative samples is set to 99 and the average user interaction is less than 9 in our datasets, an embedding matrix (512 * 512) with a maximum incremental number of 512 is sufficient. Even if a user's interaction count exceeds 512, we only need to expand the whole-word embedding matrix, which is acceptable for a real-world deployment.


\end{document}